\pdfoutput=1

\documentclass[11pt]{article}

\usepackage{ACL2023}

\usepackage{times}
\usepackage{latexsym}

\usepackage[T1]{fontenc}

\usepackage[utf8]{inputenc}

\usepackage{microtype}

\usepackage{inconsolata}

\usepackage{amsthm,amsmath,amssymb,amsfonts}
\usepackage{booktabs, multirow, multicol}
\usepackage{graphicx}
\usepackage{color, colortbl}
\usepackage{tikz}
\usepackage{soul}
\definecolor{lowred}{rgb}{1,0.80,0.80}
\definecolor{darkred}{rgb}{0.6,0.0,0.0}
\definecolor{grey}{rgb}{0.90,0.90,0.90}

\makeatletter
\newcommand*\bdot{\mathpalette\bdot@{.5}}
\newcommand*\bdot@[2]{\mathbin{\vcenter{\hbox{\scalebox{#2}{$\m@th#1\bullet$}}}}}
\makeatother
\newcommand{\neigh}[3]{\mathcal{N}_{#1}^{(#2)}\left(#3\right)}
\newcommand{\kneigh}[2]{\neigh{k}{#1}{#2}}
\newcommand{\jaccdist}[2]{1-\frac{\vert #1 \cap #2 \vert}{\vert #1 \cup #2 \vert}}

\newcommand{\vocab}{W}
\newcommand{\vectorspaceT}{V^{(T)}}
\newcommand{\vectorspaceOne}{V^{(T1)}}
\newcommand{\vectorspaceTwo}{V^{(T2)}}
\newcommand{\reprVector}[2]{\mathbf{#1}^{(#2)}}
\newcommand{\reprVectorT}[1]{\reprVector{#1}{T}}
\newcommand{\wTvector}{\reprVectorT{w}}
\newcommand{\synonymsSet}{\mathcal{S}}
\newcommand{\synOne}{u}
\newcommand{\synTwo}{v}
\newcommand{\synpair}{(\synOne,\synTwo)}
\newcommand{\firstTime}{T1}
\newcommand{\secondTime}{T2}

\title{A Tale of Two Laws of Semantic Change: \\Predicting Synonym Changes with Distributional Semantic Models}

\author{  Bastien Li\'etard \and Mikaela Keller \and
        Pascal Denis \\
        Univ. Lille, Inria, CNRS, Centrale Lille, UMR 9189 - CRIStAL, F-59000 Lille, France \\ \texttt{first\_name.last\_name@inria.fr}}

\begin{document}
\maketitle
\begin{abstract}
Lexical Semantic Change is the study of how the meaning of words evolves through time. 
Another related question is whether and how lexical relations over pairs of words, such as synonymy, change over time. 
There are currently two competing, apparently opposite hypotheses in the historical linguistic literature regarding how synonymous words evolve: 
the Law of Differentiation (LD) argues that synonyms tend to take on different meanings over time, whereas the Law of Parallel Change (LPC) claims that synonyms tend to undergo the same semantic change and therefore remain synonyms. 
So far, there has been little research using distributional models to assess to what extent these laws apply on historical corpora.
In this work, we take a first step toward detecting whether LD or LPC operates for given word pairs. After recasting the problem into a more tractable task, we combine two linguistic resources to propose the first complete evaluation framework on this problem and provide empirical evidence in favor of a dominance of LD. We then propose various computational approaches to the problem using Distributional Semantic Models and grounded in recent literature on Lexical Semantic Change detection. Our best approaches achieve a balanced accuracy above $0.6$ on our dataset. We discuss challenges still faced by these approaches, such as polysemy or the potential confusion between synonymy and hypernymy.

\end{abstract}

\section{Introduction}

Recent years have seen a surge to model lexical semantic change (LSC) with computational approaches based on Distributional Semantic Models (DSMs) \citep{tahmasebi-etal-2021-survey}.
While most research in this area has concentrated on developing approaches for automatically \textit{detecting} LSC for individual words, as in the dedicated SemEval 2020 shared task \citep{schlechtweg-etal-2020-semeval}, there has also been some work on validating or even proposing laws of semantic changes through new DSM-based approaches \citep{dubossarsky-etal-2015-bottom-up, hamilton-etal-2016-diachronic,dubossarsky-etal-2017-outta}. Ultimately, this line of work is very promising as it can provide direct contributions to the field of historical linguistics.

In this paper, we consider two laws of semantic change that are very prominent in historical linguistics, but that have to date given rise to very little computational modeling studies. Specifically, the Law of Differentiation (LD), originally due to \citet[chapter~2]{breal-1897-essai}, posits that
synonymous words tend to take on different meanings over time; or one of them will simply disappear.\footnote{To cite Bréal \shortcite{breal-1897-essai}: \textit{``[S]ynonyms do not exist for long: either they differ, or one of the two terms disappears.''}} The same idea is also discussed in more recent work, such as \citet{clark_1993}. As an example, the verbs \textit{spread} and \textit{broadcast} used to be synonyms (especially in farming), but now the latter is only used in the sense of \textit{transmit}, by means of radio, television or internet. The verbs \textit{plead} and \textit{beseech} are synonyms, but \textit{beseech} is no longer used nowadays compared to \textit{plead}. By contrast, the Law of
Parallel Change (LPC),\footnote{Name coined by \citet{xu-kemp-2015-evaluation}.} inspired from the work of \citet{stern-1921-swift}, claims that two synonyms tend to undergo the
same semantic change and therefore remain synonyms. As an illustration, \citet[chapter~3 and 4]{stern-1921-swift} describes the change of \textit{swiftly} and its synonyms from the sense of \textit{rapidly} to the stronger sense of \textit{immediately}. \citet{Lehrer-1985-influence} also observes a parallel change affecting animal terms which acquire a metaphorical sense.

These two laws are interesting under several aspects. Firstly, these laws go beyond the problem of detecting semantic change in individual words, as they concern the question of whether a lexical relationship between words, in this case synonymy, is preserved or not through time. Secondly, these laws make very strong, seemingly opposite, predictions on how synonyms evolve: either their meanings diverge (under LD) or they remain close (under LPC). It is likely that both of these laws might be at work, but they possibly apply to different word classes, correspond to different linguistic or extra-linguistic factors, or operate at different time scales. 
A large-scale study, fueled by computational methods over large quantities of texts, would be amenable to statistical analyses addressing these questions. In this work, we focus on predicting the persistence (or disappearance) of synonymy through time, as a first step toward more complete analyses.

Prima facie, DSMs appear to provide a natural resource for constructing a computational approach for assessing the importance of these laws, as they inherently --through the distributional hypothesis-- capture a notion of semantic proximity, which can be used as a proxy for synonymy. Following this idea, \citet{xu-kemp-2015-evaluation} propose the first DSM-based method for predicting how synonymous word pairs of English evolve over time (specifically, from 1890 to 1990). This research decisively concludes that there is "evidence against the Law of Differentiation and in favor of the Law of Parallel Change" for adjectives, nouns and verbs alike (i.e., the three considered POS). However, this pioneering work suffers from some limitations that cast some doubts on this conclusion. First off, the predictions made by their approach are not checked against a ground truth, thus lacks a proper evaluation. Second, the approach is strongly biased against LD, as only pairs in which \textit{both} words have changed are considered, excluding pairs in which differentiation may occur (e.g. in \textit{spread}/\textit{broadcast}, only the latter word changed in meaning).

This paper addresses these shortcomings by introducing a more rigorous evaluation framework for testing these two laws and evaluating computational approaches. We build a dataset of English synonyms that was obtained by combining lexical resources for two time stamps (1890 and 1990) that records, for a given list of synonym pairs at time 1890, whether these pairs are still synonymous or not in 1990. The analysis of this dataset reveals that, contra \citet{xu-kemp-2015-evaluation} and though using the same initial synonym set, synonymous words show a strong tendency to differentiate in meaning over time. With some variation across POS, we found that between $55$ and $80\%$ of synonyms in 1890 are no longer synonyms in 1990.  

Moreover, we propose several new computational approaches\footnote{The code used to run experiments in this paper can be found at \url{https://github.com/blietard/synonyms-semchange}}, grounded in more recent DSMs, for automatically predicting whether synonymous words diverge or remain close in meaning over time, which we recast as a binary classification problem. Inspired by Xu \& Kemp \shortcite{xu-kemp-2015-evaluation}, our first approach is unsupervised and tracks pairwise synchronic distances over time, computed over SGNS-based vector representations. Our second approach is supervised and integrates additional variables into a logistic regression model. This latter model achieves a balanced accuracy above $0.6$ over the proposed dataset.

\section{Related Work}

Data-driven methods to detect LSC have gained popularity in the recent years \cite{tahmasebi-etal-2021-survey}, using increasingly powerful and expressive word representations, ranging from the simple co-occurrence word vectors \citep{sagi-etal-2011} to static word embeddings \citep{schlechtweg-etal-2019-wind} and transformer-based contextualized word representations \citep{kutuzov-2022-contextualized,fourrier-montariol-2022-caveats}. This line of research lead to the development of shared tasks \citep{d-zamora-reina-etal-2022-black, schlechtweg-etal-2020-semeval, rodina-kutuzov-2020-rusemshift}. Most often, these tasks concern the evolution of individual words, in effect focusing on \textit{absolute} semantic change (of words individually). In this paper, we take a different stand, considering the problem of \textit{relative} change in meaning among pairs of words, specifically focusing on synonym pairs.

Previous work on word pairs are rare in the current LSC research landscape. A first exception is \cite{turney-etal-2019-natural-selection}, who also study the evolution of synonyms. They propose a dataset to track how usage frequency of words evolve over time within a sets of synonyms, as well as a new task: namely, to predict whether the dominant (most frequent) word of a synonyms set will change or not. This task is actually complementary to the one we address in this work. While \citet{turney-etal-2019-natural-selection} assume the stability of most synonym pairs between 1800 and 2000, and rather investigate the dynamic inside sets of synonymous words across time, we question this alleged stability and attempt to track whether these words remain synonymous at all in this time period.

Another distinctive motivation of our work is in the empirical, large-scale evaluation of two proposed laws of semantic change, originating from historical linguistics. Previous work investigating laws of semantic change with DSMs include \citet{dubossarsky-etal-2015-bottom-up} and \citet{hamilton-etal-2016-diachronic}, who measured semantic change of words between 1800 and 2000 and attempted to draw statistical laws of semantic change from their observations. Later, \citet{dubossarsky-etal-2017-outta} contrasted these observations and showed that even if these effects may be real, it may be to a lesser extent.

The closest work to the current research is the study of \citet{xu-kemp-2015-evaluation}, as they already focus on the two laws of Differentiation (LD) and Parallel Change (LPC). Their main motivation was to automatically measure, using DSMs, which of the two laws was predominant between 1890 and 1999. To study which of the two laws actually operates, they focus on word pairs that (i) are synonyms in the 1890s and (ii) where both words changed significantly in meaning between 1890 and the 1990s. First, they represent words as probability distributions of direct contexts, using normalized co-occurrence count vectors. Then, they measure the (synchronic) semantic proximity of words by computing the Jensen-Shannon Divergence between the corresponding distributions. Semantic change in a word is quantified by comparing its semantic space neighborhoods in the 1890s and in the 1990s. Finally, for every selected synonymous pair, they pick a control word pair that has a smaller divergence in the 1890s than the associated synonyms. At a later time in the 1990s, if the divergence for the synonyms is larger than that for the control pair, they decide these synonyms have undergone LD, otherwise they predict LPC. Ultimately, they found that most pairs (around $60\%$) have undergone LPC, which would be the dominant law.

The pioneering work of \citet{xu-kemp-2015-evaluation} faces a number of shortcomings. First, their restriction to synonymous pairs in which both words changed mechanically excludes certain cases of LD (i.e., where one one word has changed), thus introducing an artificial bias against LD. Moreover, they often select near-synonyms as controls (e.g. \textit{instructive} and \textit{interesting}) because they constrain control pairs to be \textit{closer} in divergence in the 1890s than the associated synonym pairs. Furthermore, and more importantly, \citet{xu-kemp-2015-evaluation} did not compare their predictions to any ground-truth and there is no evaluation of the reliability of their method. Finally, their choice of word representations is not among the State-of-the-Art for static methods.

In this paper, we consider all synonymous pairs, thus avoiding the bias against LD. We propose different approaches that we compare to \citet{xu-kemp-2015-evaluation}'s control pairs, and we provide results obtained with more recent distributional semantic models. Most importantly, we propose a complete evaluation framework to benchmark the different methods, something missing in this prior work.

\section{Problem Statement}
\label{ProblemStatement}

Our overarching goal is to develop new computational approaches that are able to automatically predict which pairs of synonymous words underwent LD or LPC. These predictions could be used as a first step towards providing a more refined and statistically meaningful analysis of the two laws. An important milestone towards developing such an approach is to compare it to some ground truth. Otherwise, there is no way to assess whether statistics obtained for LD or LPC are indeed reliable, a problem faced by \citet{xu-kemp-2015-evaluation}.

Unfortunately, there is no existing large-scale resource that records instances of LD/LPC, beyond a handful of examples found in research papers and textbooks in historical linguistics. What exists however are historical lists of synonyms, which we can compare to obtain some form of ground truth. This forces us to consider a slightly different methodological framework, focusing on a more constrained prediction task, namely to detect pairs of synonyms at time $T1$ that have remained synonymous or that are no longer synonymous at time $T2 (>T1)$. 

\subsection{Formalization}
\label{sec:formalization}

Let us denote $\vocab^{(T)}$ the set of words (or vocabulary) for a given language (say English) at time $T$. As language evolves through time, vocabularies at two times $T1$ and $T2$ 
need not have the exact same extensions: e.g., a word $w$ in $\vocab^{(T1)}$ might not be in $\vocab^{(T2)}$ (i.e., $w$ has disappeared). Making a simplistic, idealized assumption, let $\mathcal{C}$ be a mostly atemporal and exhaustive discrete set of concepts, and denote $M_w^{(T)}\subset \mathcal{C}$ the meaning of word $w$ at time $T$. The definition of $M_w^{(T)}$ as a set allows homonymy and/or polysemy to be accounted for. 

Given these notations, we have that $u\in\vocab^{(T)}$ and $v\in\vocab^{(T)}$ are synonyms at a time $T$ if $M_u^{(T)}\cap M_v^{(T)}\neq\emptyset$. We understand that the study of LD / LPC implies to track (i) the change of $M_u^{(T)}$ and $M_v^{(T)}$ over time, (ii) the evolution of $M_u^{(T)} \cap M_v^{(T)}$ and (iii) the very persistence of both words in vocabularies $\vocab^{(T)}$ between $\firstTime$ and $\secondTime$. Discussion about formalizing LD and LPC under those conditions can be found in appendix \ref{appendix:formalizing}.

 \subsection{Task Formulation: Tracking Synonyms Change}
\label{sec:classif}

The presented formulation, though very idealized, should make it clear that the development of a computational system that attempts to directly predict LD and LPC, and even the construction of an evaluation benchmark for evaluating such a system, are very challenging tasks. First, the initial synonym set selection presupposes, not only that one has access to a list of synonyms at $T1$ and $T2$, but also that one can reliably predict LSC in one of the two words from $T1$ to $T2$; unfortunately, LSC is still an open problem for current NLP models. Second, one typically does not have meaning inventories or automatic systems (e.g. WSD systems) for mapping words to their meanings at different time stamps. Finally, even tracking the disappearance of words through time is not trivial, as it ideally requires full dictionaries at different time stamps. 

Given these limitations, we suggest to narrow down our target problem to the task of predicting, for a given pair of synonymous words $(u,v)$ at $T1$, whether $(u,v)$ are still synonymous or not at $T2$. Stated a little more formally, we are concerned with the following binary classification problem:
\begin{align*}   
    f: & \synonymsSet^{(T1)} \rightarrow \{\text{"Syn"},\text{"Diff"}\} \\
       & (u,v)              \mapsto f( (u,v) ) = \begin{cases} \text{"Syn"~if~ $(u,v)\in\synonymsSet^{(T2)}$} \\  \text{"Diff"~otherwise} \end{cases}   
\end{align*}
where $\synonymsSet^{(T)}$ is a set of synonymous word pairs at time $T$, "Syn" indicates that words $(u,v)$ that were synonymous at $T1$ remain synonymous at $T2$, while "Diff" signals that they are no longer synonymous at $T2$. This simpler problem leads to a more operational evaluation procedure, which does not require access to $M_u^{(T^*)}$ and $M_v^{(T^*)}$, but only to lists of synonyms $\synonymsSet^{(T1)}$ and $\synonymsSet^{(T2)}$. See Section~\ref{sec:dataset} for presentation of such procedure. It should be clear that predicting which synonym pairs remain ("Syn") or cease to be synymoms ("Diff"), will provide some information about LPC and LD, although the mapping between the two problems is not one-to-one. Even if ``Diff'' covers pretty well LD, a pair that is still synonymous at $T2$ could either be a case of LPC (their shared meaning changed the same way for both words) or a pair of words that simply have not changed in meaning at all (or at least that their shared meaning is unchanged).

Now turning to designing a computational system that detects "Syn" vs. "Diff", a natural question that emerges is whether current DSMs, commonly used for detecting LSC in individual words, are able to capture synonym changes. 
More specifically, our main hypothesis will be that one can reliably track the evolution of synonymous pairs through their word vector representations at $T1$ and $T2$. 

This approach will be instantiated into different unsupervised and supervised models in Section~\ref{sec:approaches}.

\section{Evaluation Dataset}
\label{sec:dataset}

This section presents a dataset designed to track the evolution of English synonymous word pairs between two time stamps $T1$ and $T2$, with $T2>T1$. Specifically, the two time periods considered are the 1890's decade ($T1$) and the 1990's decade ($T2$). 
For extracting synonymous pairs in the 1890's (noted $\synonymsSet^{(T1)}$), we use Fernald's \textit{English Synonyms and Antonyms} \citep{fernald1896english} as \citet{xu-kemp-2015-evaluation} did. Pairs were selected based on a set of specific target words (see appendix \ref{appendix:targets}). As shown in Table~\ref{datadescr}, we obtain $1,507$ adjective pairs, $2,689$ noun pairs and $1,489$ verb pairs. 
To assess whether these word pairs are still synonyms in the 1990's, we use WordNet \citep{fellbaum2010wordnet}, as this lexical database  was originally constructed in 1990's. Thus, WordNet provides us with $\synonymsSet^{(T2)}$. Specifically, we considered that a pair of words/lemmas $(u,v)\in\synonymsSet^{(T1)}$ are still synonymous if they point to at least one common \textit{synset} in WordNet. 

\begin{table}[ht]
\centering
\small
\begin{tabular}{r|cccc}\toprule
     Synonyms pairs &ADJ &NN &VERB &All \\
     \cmidrule{1-5}
     Synonyms at $T1$ & 1507 &2689 & 1489 &5685\\\cmidrule{1-5}
     \& synonyms at $T2$ & 202 & 347 & 311 & 860\\
     \& synonyms at $T2$(\%) & 13.4 & 12.9 & 20.9 & 15.1\\
     \cmidrule{1-5}
     \& hypernyms at $T2$ & 0 & 858 & 398 & 1256\\
     \& hypernyms at $T2$(\%) & 0.0 & 31.9 & 26.7 & 22.1\\
     \& hyp. at $T2$ (1) (\%) & 0.0 & 23.2 & 22.5 & 16.9\\
     \& hyp. at $T2$ (2) (\%) & 0.0 & 6.9 & 3.5 & 4.1\\
     \& hyp. at $T2$ (3) (\%) & 0.0 & 1.4 & 0.5 & 0.8\\
     \bottomrule
\end{tabular}
\caption{\label{datadescr} Numbers of synonymous pairs extracted from \citet{fernald1896english} ($T1$) displayed by POS, and numbers of those that are also considered as synonyms or hypernyms/hyponyms in WordNet ($T2$) For hypernyms, we detail the proportions of hypernym/hyponym pairs that are separated by 1, 2 or 3 nodes in the WordNet graph.}
\end{table}

The construction of this dataset relies on two crucial hypotheses, which seem reasonable to make. First, both lexical resources rely on the same definition of synonymy. Second, $\synonymsSet^{(T2)}$ meets some exhaustivity criterion, in the sense that $(u,v)\in\synonymsSet^{(T1)}$ not appearing in $\synonymsSet^{(T2)}$ should indicate that $u$ and $v$ are no longer synonymous at $T2$, and not be due to a lack of coverage of the resource (i.e., a false negative). WordNet is assumed to be exhaustive enough, as we checked that every word involved in at least one synonymous pair has its own entry in WordNet's database.

Table~\ref{datadescr} provides some detailed statistics on the evolution of synynomous pairs between decades 1890's and 1990's, overall and for different parts of speech. A first observation on these datasets is that the proportion of pairs that are still synonyms at $T2$ (``\textit{Syn}'') is globally $15.1$\%. This implies that most synonymous pairs underwent differentiation. While it does not provide information about how change happened between $T1$ and $T2$ for the remaining $84.9$\%, it's a clue that the Law of Differentiation should be a dominant phenomenon among synonyms. 

We exploit the structure of the WordNet database to analyze the different cases of ``\textit{Diff}''. WordNet includes lexical relations of hyper-/hypo-nymy (e.g., \textit{seat}/\textit{bench}) as well as holo-/mero-nymy (e.g., \textit{bike}/\textit{wheel}) and antonymy (e.g., \textit{small}/\textit{large}) defined over synsets\footnote{As we did for synonyms, we assume that two words $w_1$ and $w_2$ are instances of one of these relations $R$ if $R$ holds for one of their corresponding synset pair.}. Note that the hyper-/hypo-nymy relation does not exist in WordNet among adjectives.   
Among nouns and verbs, we observe that around 30\% of pairs that were synonyms at $T1$ are in an hyper-/hypo-nymy relation at $T2$ and two third of them are direct hypernyms in WordNet (their synsets are direct parent/child) indicating the preservation of a very close semantic link. For a further depiction of the dataset in terms of distance in WordNet's graph, see Figure \ref{fig:wndistances} in appendix \ref{appendix:WNdistances}.

One cannot entirely exclude that $\synonymsSet^{(T1)}$ includes some hyper-/hypo-nyms as synonyms. However, even if we extend the notion of synonymy at $T2$ to include these cases, we would have only around $45\%$ of all pairs still considered synonyms among nouns and verbs. This indicates that "Diff" largely remains the most common phenomenon with an estimated proportion between $55\%$ and $80\%$. This finding contradicts the experimental results reported by \citet{xu-kemp-2015-evaluation} with their computational approach (only $40\%$ of differentiation). 

In lack of additional indication that some of these hyper-/hypo-nym cases at $T2$ are indeed synonyms, or that they may also have been hyper-/hypo-nym at $T1$, we decided to still consider them as instances of ``Diff''. Another argument for this decision is precisely that there are well-known reported cases of lexical semantic changes in which the meaning of a particular word in effect "widens" to denote a larger subset (i.e., becomes an hypernym): this is the case of \textit{dog} in English that used to denote a specific breed of dogs \citep{traugott_dasher_2001}.

\section{Approaches}
\label{sec:approaches}
This section presents two classes of computational approaches, unsupervised and supervised, for predicting whether pairs of synonyms at $\firstTime$ remain synonyms ("Syn") or cease to be so ("Diff") at a later time $\secondTime$. Common to all of these approaches is that they are based on two time-aware DSMs, one for each time stamp.

\subsection{Time-aware DSMs}

Inspired by work on LSC, we rely on separate DSMs for each time stamp $\firstTime$ and $\secondTime$, respectively yielding vector spaces $\vectorspaceOne$ and $\vectorspaceTwo$ encoding the (possibly changing) word meanings at $\firstTime$ and $\secondTime$. Thus, for each synonym pair $\synpair$, we have two pairs of vectors~: $(\reprVector{\synOne}{\firstTime},\reprVector{\synTwo}{\firstTime})\in\vectorspaceOne\times\vectorspaceOne$ and $(\reprVector{\synOne}{\secondTime},\reprVector{\synTwo}{\secondTime})\in\vectorspaceTwo\times\vectorspaceTwo$.

Specifically, we use pre-computed SGNS \citep{mikolov-etal-2013-sgns} from \citet{hamilton-etal-2016-diachronic} trained on the \textit{English} part of the GoogleBooks Ngrams dataset\footnote{\url{https://storage.googleapis.com/books/ngrams/books/datasetsv3.html}} for every decade between 1800 and 2000 and extract $\vectorspaceOne$ (1890) and $\vectorspaceTwo$ (1990). For any word $w\in\vocab$ and any time period $T$, $\wTvector \in \vectorspaceT$ is a single 300 dimensional vector. 
We ensure synonymy is accurately reflected by checking that synonym pairs have a smaller cosine distance than non-synonymous pairs for both time periods, as in Figure \ref{fig:sanityCheckSynonymy} of appendix \ref{appendix:checkSyn}.

Traditional DSM-based approaches for detecting LSC are based on self-similarities over time for a given word. For instance, for a given time interval $(T1,T2)$, they compute for each word $w$ an individual \textit{Diachronic Distance}, noted here $DD^{(T1,T2)}(w)$. Cosine distance is often used (recall in appendix \ref{appendix:definitions}).

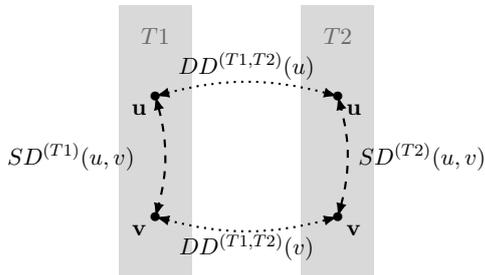
\begin{figure}[ht]
    \centering
    \begin{tikzpicture}[thick,scale=0.8, every node/.style={scale=0.8}]
        \coordinate (v_O) at (0, 0);
        \coordinate (u_O) at (0, 2);
        \coordinate (v_E) at (3, 0);
        \coordinate (u_E) at (3, 2);
        \coordinate (textDDu) at (1.5, 2.5);
        \coordinate (textDDv) at (1.5, -.5);
        \coordinate (textSDE) at (3.2, 1);
        \coordinate (textSDO) at (-.2, 1);
        
        \fill [color=gray!30] (u_O)++(-.6,1.5) rectangle (0.6,-1);
        \fill [color=gray!30] (u_E)++(-.6,1.5) rectangle (3.6,-1);
        \draw (u_O)++(0, 1) node[color=black!60] {$\firstTime$};
        \draw (u_E)++(0, 1) node[color=black!60] {$\secondTime$};
        \draw (v_O) node {$\bullet$} ;
        \draw (v_O) node[below left]{$\mathbf{\synTwo}$};
        \draw (v_E) node {$\bullet$} ;
        \draw (v_E) node[below right]{$\mathbf{\synTwo}$};
        \draw (u_O) node {$\bullet$} ;
        \draw (u_O) node[below left]{$\mathbf{\synOne}$};
        \draw (u_E) node {$\bullet$} ;
        \draw (u_E) node[below right]{$\mathbf{\synOne}$};

        \draw [latex-latex, dotted] (u_O) to [bend left=15] (u_E);
        \draw (textDDu) node {$DD^{(\firstTime,\secondTime)}(\synOne)$};
        
        \draw [latex-latex, dashed] (u_O) to [bend left=15] (v_O);
        \draw (textSDO) node[left] {$SD^{(\firstTime)}\synpair$};
        
        \draw [latex-latex, dotted] (v_O) to [bend right=15] (v_E); 
        \draw (textDDv) node {$DD^{(\firstTime,\secondTime)}(\synTwo)$}; 
        
        \draw [latex-latex, dashed] (v_E) to [bend right=15] (u_E); 
        \draw (textSDE) node[right] {$SD^{(\secondTime)}\synpair$};

    \end{tikzpicture}
    \caption{Pairs of word embeddings at 2 time periods and associated diachronic and synchronic distances.}
    \label{fig:vectorsAndMeasures}
\end{figure}

There is no obvious distance for comparing \textit{pairs} of word vectors, but one can instead rely on comparing the pairwise word vector distance at each time stamp $T$; we call this \textit{Synchronic Distance} (denoted SD). The two types of distances for two time stamps $\firstTime$ and $\secondTime$ are described in Figure~\ref{fig:vectorsAndMeasures}.  Our unsupervised method, proposed in Sec.~\ref{DivBasedMethod} directly exploit the idea of tracking different types of SD through time, while Sec.~\ref{supervised} presents a supervised approach that combines both SD and DD.

\subsection{Unsupervised Methods}
\label{DivBasedMethod}

While we don't have access to $M_{u}^{(T)}$ and $M_{v}^{(T)}$, we can represent the meaning of $u$ and $v$ using DSM and compare them at a given time to estimate how close they are in meaning. Indeed, if $M_{u}^{(T)} \cap M_{v}^{(T)}$ changes, this should be reflected in difference of the use contexts of $u$ and those of $v$, and so reflected in the distance between $\reprVectorT{\synOne}$ and $\reprVectorT{\synTwo}$. 
Let 
\begin{equation*}
    SD^{(T)}: \vocab^{(T)} \times \vocab^{(T)} \rightarrow \mathbb{R}^+
\end{equation*}
 be a measure of \textbf{synchronic distance} between vectors representing two words. By construction of $\vectorspaceT$, $SD^{(T)}\synpair$ is smaller for words $\synpair$ that appear in similar contexts than for unrelated words. We assume that there exists a value $\delta_T$ such that 
 \begin{equation*}
 \forall (u,v)\in\synonymsSet^{(T)},\> SD^{(T)}(u,v) \le \delta_T.
 \end{equation*}
 This entails that for a given pair $(u,v)$: 
 \begin{equation*}
 SD^{(T)}(u,v) > \delta_T \Rightarrow (u, v) \textrm{ are not synonyms}.
 \end{equation*}
In this setting, one can compare the synchronic distances within $\vectorspaceOne$ and with $\vectorspaceTwo$ and decide if the pair differentiated or stayed synonymous.

Let $\synpair$ be a pair of synonyms at $\firstTime$, as such we have that $
SD^{(\firstTime)}(u,v)\le\delta_{\firstTime}$.
If $(u,v)$ are not synonyms at time $\secondTime$ then 
$SD^{(\secondTime)}(u,v)>\delta_{\secondTime}$.

Combining these two inequalities, we would say that a pair of synonyms at $\firstTime$ has differentiated at $\secondTime$ if: 
\begin{equation*}
\underbrace{SD^{(\secondTime)}\synpair-SD^{(\firstTime)}\synpair}_{\textstyle\begin{gathered} =\Delta\synpair \end{gathered} } >\delta_{\secondTime}-\delta_{\firstTime}.
\end{equation*}
Ideally one could imagine that the distance threshold $\delta_T$ at which, words cease to be synonyms should be independent of the time period $T$. Empirically however, because word embeddings are not necessarily build with an enforced scale, there might be a dilation or shrinking in the overall synchronic distances between $\firstTime$ and $\secondTime$. Let us assume that 
\begin{equation*}
\delta_{\secondTime}=\delta_{\firstTime}+\tau,\>\tau\in\mathbb{R}.
\end{equation*}  

Our decision rule could then be rewritten as:
\begin{equation}
\label{divCondition}
f(u,v) = \begin{cases}
    \text {``Diff''  if } \Delta(u,v) \geq \tau\\
    \text {``Syns'' otherwise.}
\end{cases}
\end{equation}

This approach is shortly denoted ``$\Delta$'' in section \ref{sec:experiments}. It diverges from the prior work of \citet{xu-kemp-2015-evaluation} that chooses to rely on control pairs instead of a threshold. For the sake of comparison, we implemented their method presented as ``\textit{XK controls}''. It is not the full protocol presented by \citet{xu-kemp-2015-evaluation}, as (i) the experimental setting is not identical, they filtered out some synonym pairs and we didn't (ii) we use SGNS word representations and cosine distance instead of normalized co-occurrence counts and Jensen-Shannon Divergence. \citet{schlechtweg-etal-2019-wind} provided a longer comparison between word representations.

We propose a statistically-grounded criterion to set the value for the threshold $\tau$. Since the meaning of most words is expected to remain stable\footnote{Intuitively, someone in 2023 can still understand writings published in the 1890s in their original text, like books from Charles Dickens or Arthur Conan Doyle.}, we argue that most pairwise distances should remain stable as well. We can then estimate the dilation between the representations in the two time periods by the average gap between the synchronic distances of words. 

\begin{equation}\tau = \frac{1}{|{\vocab}|^2}\sum_{(w_1,w_2)\in{\vocab}\times{\vocab}} \Delta(w_1,w_2)\end{equation}

In practice, we experiment with two different types of synchronic distances between words. The first is the cosine distance (see \ref{appendix:definitions}). That is: 
\begin{equation*}
    SD^{(T)}(u,v) = 
    \operatorname{cos-dist}(\reprVectorT{u},\reprVectorT{v}).
\end{equation*}
We shortly denote it ``SD(cd)''. Another measure of semantic proximity is based on the shared word neighborhood between the two vectors $u$ and $v$:
\begin{equation*}
    SD^{(T)}(u,v) = \operatorname{jaccard-dist}(\kneigh{T}{u}, \kneigh{T}{v}),
\end{equation*} with $\kneigh{T}{w}$ being the set of the $k$-nearest neighbors of the point representing $w$ in the vector space at time $T$, and \textit{jaccard-dist} being the Jaccard distance (see appendix \ref{appendix:definitions}). This measure is ranged between 0 and 1, and we denote it ``SD(n$k$)''.

\subsection{Supervised Methods}
\label{supervised}
Approaches described so far use the labels in the dataset ("Syn" and "Diff") only for evaluation purposes. But one can also use part of the available data to learn a \textit{supervised} classifier to predicts these labels.  Concretely, for most of these models, we trained Logistic Regression (LR) models\footnote{Implemented with the \textit{scikit-learn} library for Python\footnote{\url{https://scikit-learn.org/}}.}

\paragraph{Synchronic Distances Combination}
In our unsupervised approach, we compute $SD^{(\firstTime)}$ and $SD^{(\secondTime)}$ and their difference, denoted $\Delta$. This quantity is then compared to a fixed threshold $\tau$. We propose to investigate two supervised approaches stemming from this: (i) simply tune $\tau$ and (ii) use a LR model to learn the optimal weighting in the linear combination of the two distances. This latter model is called ``LR SD''.

\paragraph{Accounting for Individual Change}
Most works about computational approaches to LSC focus on detecting the change of a single word \citep{tahmasebi-etal-2021-survey}, using a diachronic distance, which we noted $DD^{(\firstTime,\secondTime)}(w)$, across time periods $\firstTime$ and $\secondTime$ for individual words $w$. 

In addition to synchronic distances, we input diachronic distances as features for a LR model. The resulting classifier (LR SD+DD) uses the 4 distances represented in Figure \ref{fig:vectorsAndMeasures} as variables: self-similarities across time periods (\textit{DD}s), and a distance measure within pairs for each of both time stamps (\textit{SD}s). Similarly to synchronic distances defined in Sec. \ref{DivBasedMethod}, we try two definitions of DD. First, we compare sets of neighbors at $\firstTime$ and $\secondTime$: 
\begin{equation*}\operatorname{DD}(w) = \operatorname{jaccard-dist}(\kneigh{\firstTime}{w},\kneigh{\secondTime}{w}).
\end{equation*} 
We also compute the cosine distance between $\reprVector{w}{\firstTime}$ and $\reprVector{w}{\secondTime}$ after aligning the vector space $\vectorspaceTwo$ to $\vectorspaceOne$ using Orthogonal Procrustes \citep{hamilton-etal-2016-diachronic, schlechtweg-etal-2019-wind, schlechtweg-etal-2020-semeval}. Denoting $\reprVector{w}{\secondTime}_{align}$ the vector $\reprVector{w}{\secondTime}$ after alignement with Orthogonal Procrustes, we have:
\begin{equation*}\operatorname{DD}(w) = \operatorname{cos-dist}(\reprVector{w}{\firstTime},\reprVector{w}{\secondTime}_{align}).
\end{equation*}

\begin{table*}[ht]
    \centering
    \begin{tabular}{r|cccc|ccc}\toprule
         \textbf{Dataset} & \textbf{ADJ} & \textbf{NN} & \textbf{VERB} & \textbf{ALL} & \multicolumn{3}{c}{\textbf{ALL}}  \\
         \textbf{Evaluation metric} & \multicolumn{4}{c|}{\textbf{Balanced Accuracy}} & $\mathbf{F_1}$\textit{(Syn)} & $\mathbf{F_1}$\textit{(Diff)} & \textbf{\%(D)}  \\
         \cmidrule{1-8}
        All (\textit{Syn}) & .50 & .50 & .50 & .50  & .48 & 0 & 0 \\
        All (\textit{Diff}) & .50 & .50 & .50 & .50  & 0 & .81 & 100 \\
        LR F & .51 & .56 & .59 & .55  & .35 & .74 & 75 \\
        \cmidrule{1-8}
        XK controls & .52 & .49 & .51 & .50 & .33 & .67 & 65\\
        $\Delta$ (cd) & .50 & .49 & .51 & .50 & .27 & .73 & 75\\
        $\Delta$ (n$k$) & .48 & .49 & .49 & .50 & .32 & .67 & 66\\
        \cmidrule{1-8}
        $\Delta$ (tuned $\tau$) & .51 & .52 & .52 & .51 & .27 & .74 & 79\\
        LR SD & .60 & .62 & .59 & .60 & .48 & .69 & 56\\
        LR SD + DD & .61 & .62 & .60 & .60  & .48 & .69 & 56 \\
        LR SD + F & .61 & \textbf{.64} & .63 & \textbf{.62}  & .51 & .71 & 57\\
        LR SD + DD + F & \textbf{.62} & \textbf{.64} & .63 & \textbf{.62}  & .50 & .70 & 57\\
        LR multi & \textbf{.62} & \textbf{.64} & \textbf{.65} & \textbf{.62}  & .51 & .71 & 57\\
        \cmidrule{1-8}
        LR multi. poly. degree (2) & .56 & .63 & .62 & \textbf{.62}  & .50 & .70 & 60\\
        SVM (gaussian) & .60 & \textbf{.64} & \textbf{.65} & \textbf{.62} & .50 & .74 & 63\\
        \bottomrule
    \end{tabular}
    \caption{Performances of the different approaches.  Results are averaged over 20 random splits.}
    \label{tab:results}
\end{table*}

\paragraph{Using Distances and Frequencies}
A final step of this process is to add word frequencies for both words at both time periods, as there exist links between usage frequency and semantic change \citet{zipf-1945-repetition}. We could observe whether adding explicit frequency information helps retrieving discriminatory clues that could be missed by using only distributional representations.

Word frequencies were estimated from the Corpus of Historical American English (COHA) list,\footnote{\url{https://www.ngrams.info/download_coha.asp}} which has the advantage to be genre-balanced. As variables for both words and both periods to feed our model, we try to add either raw occurrences counts (indicated by ``+FR''), either grouped frequency counts (``+FG''). The procedure to create such groups is described in appendix \ref{freqGroups}. 

\paragraph{All Features}
For the sake of comparison to previous models, we evaluate LR models that take as input an implementation of each of these features (SD + DD + frequency); and an even larger model (called ``LR multi.'') that reunites \textit{all} described implementations of \textit{SD}, \textit{DD} and frequencies.

\paragraph{Non-linear Models}
As a further step increasing the model's complexity, we try to combine this full set of available variables in a non-linear fashion. We compare previous models to polynomial features (degree 2) preprocessing\footnote{We also try degrees higher than 2, finding no consistent improvement.} and a SVM classifier with a Gaussian kernel.

\section{Experiments}
\label{sec:experiments}

\subsection{Experimental Settings}

\paragraph{Target Words Selection}
We use a unique vocabulary $\vocab$ composed of $6,453$ adjectives, $16,135$ nouns and $10,073$ verbs. The process to select words is described in appendix \ref{appendix:targets}.

\paragraph{Dataset Splits} For every POS tag, we have a set of word pairs that are synonymous at $\firstTime$. We call \textit{ALL} the dataset that comprises all pairs indistinctly of their POS. These datasets (ADJ,NN,VERB or ALL) are individually shuffled and 33\% of their samples (pairs) are set aside for testing. For each dataset, a model is trained on the 66\% remaining pairs and evaluated on the test part. Presented results are averaged over 20 random train/test splits.

\paragraph{Hyperparameters}
We train models with combinations of the different definitions of distances and frequency variables. Choice of synchronic distances was between SD(cd) and SD(n$k$) with $k$ in $\{5,10,15,20,40,100\}$. For $\operatorname{DD}$, we tried neighborhoods with fixed size $100$, like \citet{xu-kemp-2015-evaluation}, and Orthogonal Procrustes with cosine distances. For frequency, the choice is between raw counts and groups. The selected models are detailed in Appendix \ref{appendix:models}. The ideal value for the SVM's regularization parameter is found using 5-fold cross-validation over the training set.

\paragraph{Evaluation Metrics}
We use two standard evaluation metrics: $F_1$ score and \textit{Balanced Accuracy (BA)}. 
$F_1$ scores were computed for both classes, denoting it ``$\mathbf{F_1}$\textit{(Syn)}'' for \textit{Syns} and ``$\mathbf{F_1}$\textit{(Diff)}'' for \textit{Diff}. BA is defined as the average of recalls for both classes, and provide a notion of accuracy robust to class imbalance. We also display the percentage of predicted \textit{Diff} (``\%D'').

\paragraph{Baselines}
The first two baselines are constant output classifiers, always predicting "\textit{Syn}" or "\textit{Diff}" respectively. They are expected to have a balanced accuracy of $50\%$, as they would be fully accurate for one class and always wrong for the other. The third baseline (\textit{LR Frequency}) is a Logistic Regression model trained \textit{only} with frequency variables, without any knowledge on the semantic aspect of the pair (neither \textit{SD} or \textit{DD}).

\subsection{Results}

Performances over the test parts of the different datasets are displayed Table~\ref{tab:results}.

The first observation is that, in line with the dataset's proportions, all models predict a majority of ``Diff'', even unsupervised ones (including our reimplementation of Xu \& Kemp's control pair selection method). While our task does not directly address the question of the opposition between LD and LPC, this is an empirical clue in favor of LD, contradicting \citet{xu-kemp-2015-evaluation}. However, predicting the right amount of ``Diff'' does not guarantee the quality of predictions. Indeed, obtained balanced accuracies range between $0.49$ and $0.65$. 

Considering our unsupervised methods and the $\Delta$ (tuned $\tau$), we find no real improvement over baselines. In particular, they fail to outperform the frequency-based baseline model which performs surprisingly well. On the other hand, Logistic Regression and SVM models substantially improve over the baselines, Xu \& Kemp's control pairs and all $\Delta$-based methods. Interestingly, LR SD outperforms $\Delta$-based methods despite the fact that they rely on the same components.

The gap between baselines and models is larger for nouns and lesser for verbs. Despite these POS-specific differences, best models are consistently the ones using both SD and frequencies, while DD brings little to no improvement. This can be expected as individual changes of words seem less important on the problem of \textit{Syn}/\textit{Diff}. However, this factor could be used in future work to distinguish pairs of synonyms (among the \textit{Syn} class) that did not change and pairs that went under LPC.

We observe that there is a substantial difference in $F_1$ scores between the two classes, $\mathbf{F_1}$\textit{(Syn)} being lower than $\mathbf{F_1}$\textit{(Diff)} across all models. Moreover, models with higher $\mathbf{F_1}$\textit{(Syn)} are often found to be the ones with higher balanced accuracy, even when $\mathbf{F_1}$\textit{(Diff)} is lower. This is likely linked to the fact that the datasets are highly imbalanced as presented in Table~\ref{datadescr}: the ground truth proportion of \textit{Syn} never exceeds 21\%. We also remark that \citet{xu-kemp-2015-evaluation} decision rule based on control pairs also predicts a majority of \textit{Diff}, contrarily to the results they showed. It may be because the protocol is not fully identical.

\subsection{Confounding Factors}
\label{sec:analysisExt}

\begin{figure}
    \centering
    \includegraphics[width=0.38\textwidth]{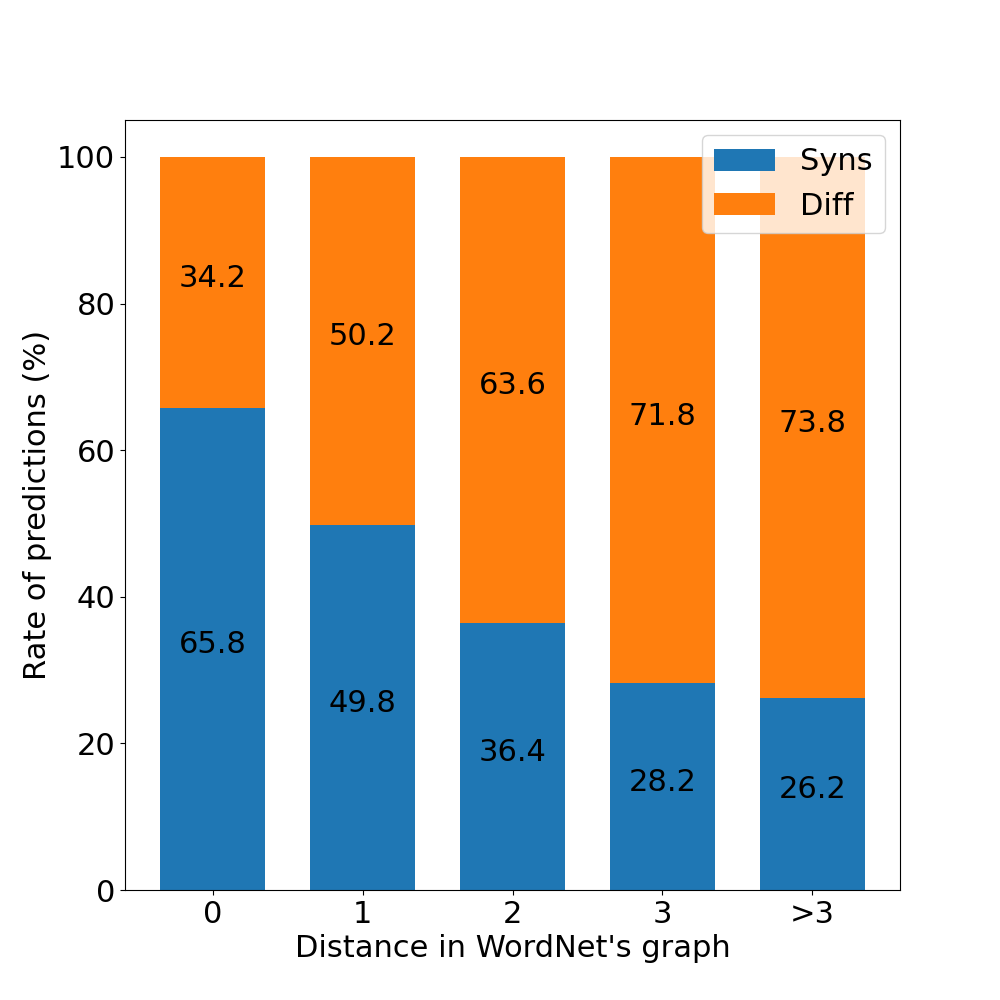}
    \caption{Proportions of predictions of the models w.r.t. the actual distance $d$ in WordNet of \textit{noun} pairs. Pairs with $d=0$ are synonymous pairs in WordNet.}
    \label{fig:WNdist_vs_preds}
\end{figure}

Using WordNet, we discuss two aspects that may be sources of errors when detecting a change in synonymy: polysemy and hypernymy. We study predictions of our best performing LR model on the noun dataset.

\paragraph{Polysemy}
WordNet provides us with different set of synonyms for every entry, corresponding to different senses or usages, and therefore we can measure the polysemy of a word at $T2$. We found that pairs misclassified as "Syn" tend to be those whose second term has fewer senses (6 senses on average as compared with well classified "Diff" which have 8 senses on average). Indeed, as we use static embeddings and no Word Sense Disambiguation (WSD) method, our model is subject to the complexity brought by polysemy. In a recent shared task about Lexical Semantic Change measures, best performing models are the one using WSD methods \citep{d-zamora-reina-etal-2022-black}. This finding highlights the importance of handling polysemy as a potential confounding factor.

\paragraph{Distances in WordNet}
In Figure \ref{fig:WNdist_vs_preds} we display the percentage of prediction with respect to shortest distance between the two words of \textit{noun} pairs in WordNet's graph. The distance $d$ is the minimum number of nodes separating the two words. 
We remark that, as expected, the model predicts more and more \textit{Diff} as $d$ increases. What is more interesting is that for $d=1$ (direct hypernymy), there is still an important proportions of predicted \textit{Syn}. This highlights that our model has difficulties to handle hypernymy and confuses it with synonymy.

\section{Conclusion}

 In this work, we considered two contradicting laws about the semantic change of synonyms. We discussed the necessary adaptations of the problem statement for this particular type of LSC and elaborated a framework to evaluate models for this new classification problem. The use of linguistic resources from two different time periods allowed us to improve model analysis with respect to prior work on the matter. Then we proposed unsupervised and supervised approaches relying on measures of semantic change extracted or inspired by existing literature on LSC, and also leveraged the usefulness of explicit word usage frequency information. We compared these approaches in our evaluation framework, finding that distances in vector spaces from different time periods should not be considered equally. We also observed that explicit frequency information actually help distributional methods to capture the change of synonymy. Finally we discussed challenges that DSM approaches still face and opened a discussion about the interplay between hypernymy and synonymy.

\section*{Limitations}

As mentioned already, the problem \textit{Syn}/\textit{Diff} does not reflect the initial question of LD/LPC. In particular, the \textit{Syn} class of pairs that remained synonyms contains pairs that underwent LPC and pairs which shared meaning remained unchanged. The latter does not play a role in the LD/LPC dichotomy and should be discarded for deeper study of the two apparently opposite laws. Also, we restrain the study to some target words that are chosen to occur at both time periods, thus preventing us to fully measure the importance of LD. Indeed, recall that Bréal's Law of Differentiation predicts that some synonyms may disappear in the process. Thus, our \textit{Diff} class could be considered incomplete. However, including such disappeared words would prevent the use of time-aware DSMs.

Section \ref{ProblemStatement} presented synonymy as a symmetrical relation between words. However, a thesaurus like \citet{fernald1896english} displays asymmetrical synonymy: for an entry $u$ we have a set of synonyms $v_1,v_2,...$ from which we extract pairs $\synpair$. We observe that $v$ itself is rarely an entry of the thesaurus, and when it does, $u$ may not appear in the list of synonyms of $v$. This is contradictory to WordNet's definition of synonymy that consider this relationship to be symmetrical. However, up to our knowledge, there is no lexical database (like WordNet) being also historical and that could help us ensure the notion of synonymy at both time periods is strictly the same. In the absence of such a resource, we leave potential disagreements in definition between the two linguistic resources to future investigations.

In section \ref{sec:dataset}, we discussed that hyper/hypo-nymy could be misleading. We made the assumption that \citet{fernald1896english} and Wordnet \citep{fellbaum2010wordnet} used similar-enough notions of synonymy such that our labels \textit{Syn}/\textit{Diff} are relevant. However, thesaurus like \citet{fernald1896english} are created as a tool for writers and authors to avoid redundancy, thus including wide lists of synonyms that include hypernyms (instead of repeating \textit{the bench}, you could say \textit{the seat}). In section \ref{sec:analysisExt} we showed that direct hypernymy is misleading for our model. Yet, we still miss guidelines/insights about the possibility to include some cases of hypernymy among synonyms at $\secondTime$. Another approach would be to remove hypernyms from the source material at $\firstTime$, which implies to automatically detect them or manually review thousands of pairs.

There are remaining factors that presented approaches do not take in account and that one could think relevant. In particular, further work could investigate the influence of pressure of words on a concept, for instance many words sharing (at least partially) a similar meaning. However, this would require access to list of senses for each word at time $\firstTime$, which we do not have in \citet{fernald1896english}. To this extent, contextualized language models fine-tuned for the different time periods could be helpful.

Finally, because we used pre-computed SGNS embeddings on historical data binned in decade, we have no guarantee that this is the optimal setting for studying Lexical Semantic Change. Maybe different kind of changes could be observed using larger or smaller time periods, and conducting the study over a larger or a smaller time span instead of just a century.

\section*{Acknowledgements}

We would like to thank the three anonymous reviewers for their helpful comments on this paper. We would also thank Anne Carlier for the thoughtful discussion about this work. This research was funded by Inria Exploratory Action COMANCHE.

\bibliography{anthology,custom}
\bibliographystyle{acl_natbib}

\appendix

\section{Appendix}
\label{sec:appendix}

\subsection{Formalizing LD and LPC}
\label{appendix:formalizing}

In this work, we reduced the problem from finding pairs in which LD or LPC operates to a binary classification problem between pairs that remained synonymous and those who did not. To understand the need for a reduction, let us introduce some notation and definitions. 

First, let us denote by $\vocab^{(T)}$ the set of words (or vocabulary) for a given language (say English) at time $T$. As language evolves through time, vocabularies at two times $T1$ and $T2$ (with $T2>T1$) need not have the exact same extensions: e.g., a word $w$ in $\vocab^{(T1)}$ might not be in $\vocab^{(T2)}$ (i.e., $w$ has disappeared) or vice versa (i.e., $w$ is a new word).  Assuming a simple, idealized denotational semantics, we will further define $\mathcal{C}^{(T)}$ as the set of discrete concepts available at time $T$,\footnote{We take $\mathcal{C}^{(T)}$ to be mostly stable over time, but new concepts might of course appear or disappear (e.g., due to techonological or cultural evolution).} and $M_w^{(T)}\subset \mathcal{C}$ the meaning of word $w$ at time $T$. It is defined as a set to model cases of homonymy and/or polysemy. From these definitions, we can now define \textit{synonymy} at time $T$ between words $u\in\vocab^{(T)}$ and $v\in\vocab^{(T)}$ as $M_u^{(T)}\cap M_v^{(T)}\neq\emptyset$; that is, $u$ and $v$ do share a common meaning. Furthermore, we can define the \textit{semantic change} from $T1$ to $T2$ in a word $w$ as follows: $M_w^{(T1)} \neq M_w^{(T2)}$; that is, $w$ has different sets of meanings at $T1$ and $T2$. 

Equipped with these definitions, we are now ready to formalize the two laws LD and LPC, starting with what their common scope.

First, both laws concern synonyms: they are restricted to a set of synonyms at some initial time $T1$, defined by $\synonymsSet^{(T1)} = \{(u,v): M_u^{(T1)}\cap M_v^{(T1)}\neq\emptyset\}$. 

Second, both LD and LPC assume some individual semantic change, from $T1$ to $T2$ (with $T2>T1$), in at least one of two synonymous words: that is, $M_u^{(T1)} \neq M_u^{(T2)}$ or (logical) $M_v^{(T1)} \neq M_v^{(T2)}$. 

Given these preconditions, the application of LD implies that either: 
\begin{itemize}
    \item one of the two words has disappeared:\\ $u\in\vocab^{(T1)} \land u\not\in\vocab^{(T2)}$ \\or (exclusive) $v\in\vocab^{(T1)} \land v\not\in\vocab^{(T2)}$, 
    \item $u$ and $v$ are no longer synonymous at $T2$:\\ $M_u^{(T1)}\cap M_v^{(T1)}=\emptyset$.
\end{itemize}

By contrast, LPC implies that words $u$ and $v$ remain synonymous from $T1$ to $T2$. While this could be simply stated as: $M_u^{(T2)}\cap M_v^{(T2)}\neq\emptyset$, we feel that this misses an important aspect of the law, namely that $M_u^{(T1)}$ and $M_v^{(T1)}$ should evolve in the same way:
\begin{itemize}
    \item either by acquiring (a) new shared sense(s): $(M_u^{(T2)}-M_u^{(T1)})\cap (M_v^{(T2)}-M_v^{(T1)})\neq\emptyset$,
    \item  or inversely by losing the same sense(s): $(M_u^{(T1)}-M_u^{(T2)})\cap( M_v^{(T1)}-M_v^{(T2)})\neq\emptyset$. 
\end{itemize}

\subsection{Useful definitions}
\label{appendix:definitions}

Recall the definition of \textit{cosine distance} between two vectors $\mathbf{x}$ and $\mathbf{y}$:
\begin{equation}
   \operatorname{cos-dist}(\mathbf{x},\mathbf{y}) = 1-\frac{\langle \mathbf{x} , \mathbf{y} \rangle}{\lVert \mathbf{x} \rVert \lVert \mathbf{y} \rVert }.
\end{equation}  

We also recall the definition of \textit{Jaccard distance} between two sets $A$ and $B$:
\begin{equation}
    \operatorname{jaccard-dist}(A,B) = \jaccdist{A}{B}.
\end{equation}

\subsection{Xu \& Kemp's control pairs}
\label{appendix:xkcontrols}

In Table~\ref{tab:controls} we display samples of word pairs selected as control pairs following \citet{xu-kemp-2015-evaluation}'s procedure. As we can observe, for every Part-Of-Speech, a significant number of these pairs are themselves synonymous. After manually reviewing a hundred pairs for each POS tag, we estimate that the proportion of synonyms in the selected control pairs is between 20 and 40\%. Synonym pairs shouldn't be used to control other synonym pairs, which may explain why our reproduction of \citet{xu-kemp-2015-evaluation} decision rule does not perform well according to Table~\ref{tab:results}. 

\begin{table*}[t]
    \centering
    \begin{tabular}{cc}\toprule
        \textbf{POS} & \textbf{Control pairs} \\\cmidrule{1-2}
        \multirow{3}{*}{ADJ} & brownish/red, kindly/mild, teeming/agricultural, likeliest/meaningless, \\
        & \textit{various/heterogeneous}, \textit{barbarous/cruel}, \textit{abandoned/unsuccessful}, trojan/escaping, \\
        & \textit{subjective/relative}, reliable/readable.\\\cmidrule{1-2}
        \multirow{3}{*}{NN} & diphtheria/typhus, \textit{muskets/pistol}, surgery/appendicitis, beech/apples,\\
        & accountants/prints, commodity/substances, \textit{cups/pots}, wife/grandmother,\\
        & fool/fisherman, obstacles/multiplication.\\\cmidrule{1-2}
        \multirow{3}{*}{VERB} & \textit{moan/groan}, divide/span, needed/secured, \textit{flowed/flooded},\\
        & stall/owned, \textit{told/asked}, \textit{mentioned/described}, cooperate/accord,\\
        & copy/filed, increased/diminished.
        \\\bottomrule
    \end{tabular}
    \caption{Random samples of size 10 among selected control pairs. In italic are control pairs which are considered synonyms according to the definition in Section~\ref{sec:formalization}.}
    \label{tab:controls}
\end{table*}

\subsection{Distances in WordNet}
\label{appendix:WNdistances}

\begin{figure}[ht]
    \centering
    \includegraphics[width=0.45\textwidth]{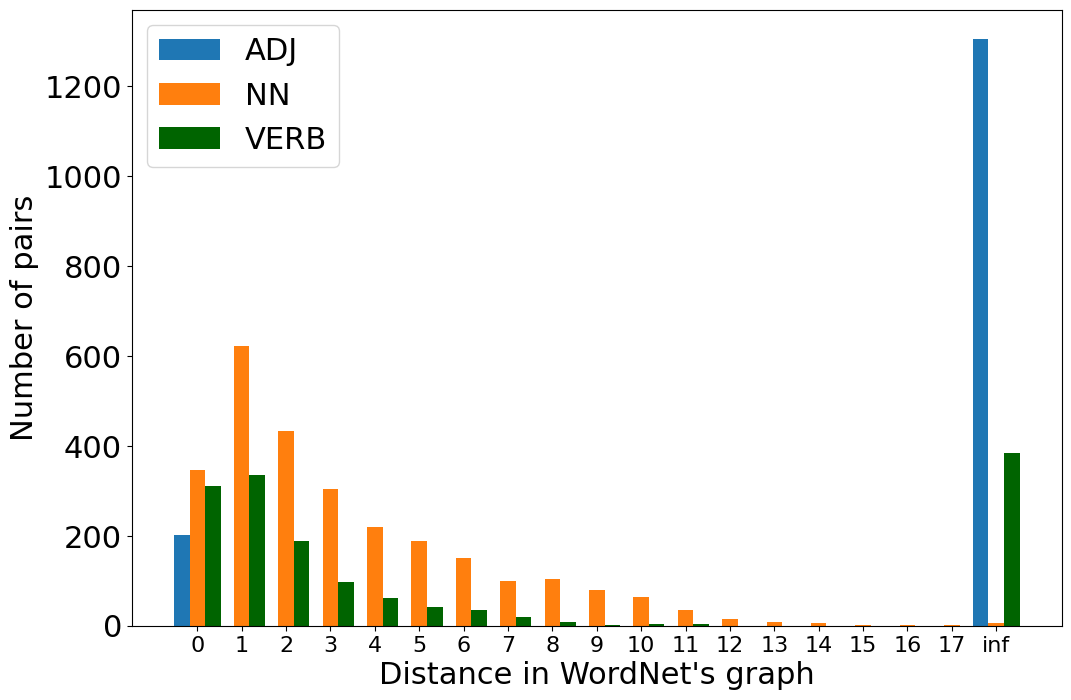}
    \caption{Distribution of shortest distances in WordNet between pairs of words that were synonymous at $T=O$. \textit{inf} means that there is no path between the two words in WN. A distance of 0 means that they are actually synonyms, while a distance of 1 implies there is direct hypernymy.}
    \label{fig:wndistances}
\end{figure}

In Figure \ref{fig:wndistances} are displayed the distributions of distances in WordNet. The distance in WordNet between two words $(u,v)$ is the number of nodes of the shortest path between a synset of $u$ and a synset of $v$.

\subsection{Synonymy in our DSMs}
\label{appendix:checkSyn}

\begin{figure}[ht]
    \centering
    \includegraphics[width=0.45\textwidth]{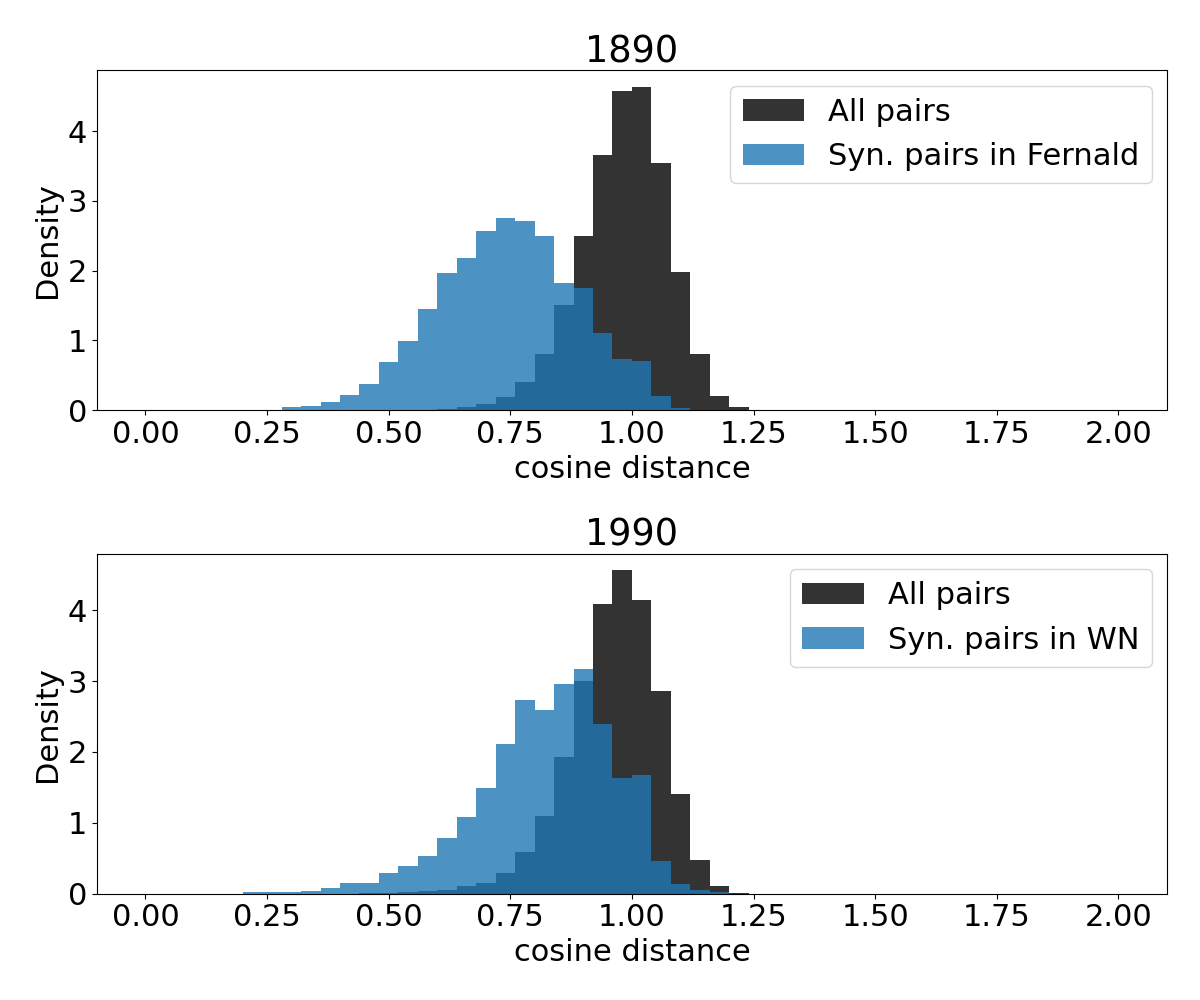}
    \caption{Distribution (as density histograms) of cosine distances between word pairs at time period $T1$ (decade 1890s) and $T2$ (decade 1990s). In blue are represented pairs of synonyms, and in black are represented all pairs of target words, without any particular constraint.}
    \label{fig:sanityCheckSynonymy}
\end{figure}

In Figure \ref{fig:sanityCheckSynonymy} are displayed the distributions of cosine distance between word pairs at both periods. In blue are synonyms at this time (from \citet{fernald1896english} at $\firstTime$, and from WordNet at $\secondTime$). In black are all possible word pairs. We observe that synonymy is indeed captured by our DSM as synonyms are significantly closer in cosine distance than other word pairs.

\subsection{Frequency groups}
\label{freqGroups}
The procedure to create a fixed number $M$ of frequency group is the following. At a time $T$, the list of target words is sorted by increasing frequency, we label as group `0' the first 50\% of the list. In the remaining 50\%, The first half is labeled as group `1', and so on until group $M-2$ is created. The still unlabeled words are labeled group $M-1$, for a total of $M$ groups. Group labels are therefore positively correlated with occurrences counts. 

\subsection{Target words selection}
\label{appendix:targets}
Among words represented in the embeddings provided by \citet{hamilton-etal-2016-diachronic}, we keep only words following these three requirements. The first is to be POS-tagged as an \textit{adjective}, a \textit{noun} and/or as a \textit{verb} in the COHA. For a given POS-tag among these three, the second requirement is to appear at least 3 times in every decade between 1890 and 1999. Lastly, we require words to be composed of 3 letters or more. If a word appears with multiple POS-tags in the COHA and fulfills the minimum frequency requirement with each of these tags, the same embedding is used as its representation, as \citet{hamilton-etal-2016-diachronic}'s training data aggregated POS-tags.

\subsection{Unsupervised models}
\label{appendix:unsup}

\begin{figure}[ht]
    \centering
    \includegraphics[width=0.5\textwidth]{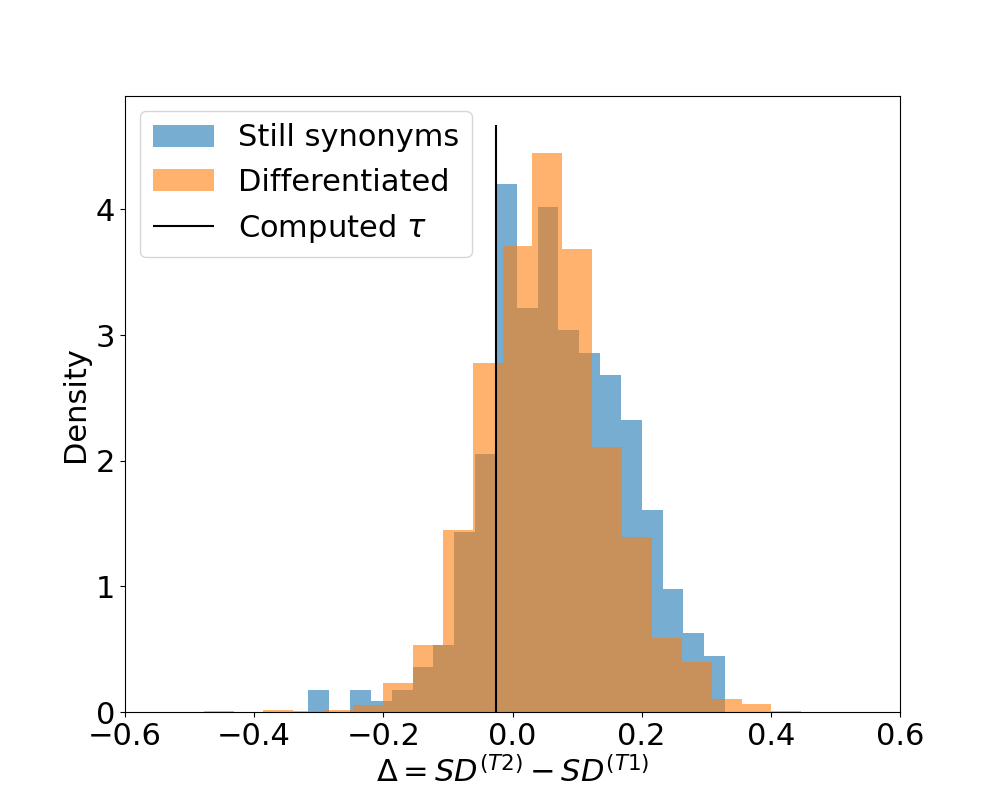}
    \caption{Histograms of the value of divergence $\Delta$ of synonymous pairs, depending whether they differentiated (orange) or stayed synonyms (blue).}
    \label{fig:divUnsupThresholds}
\end{figure}

In Figure \ref{fig:divUnsupThresholds}, we observe that the quantity $\Delta$ does not reflect a clear separation between \textit{Syn} pairs and \textit{Diff} pairs. This explains why the unsupervised methods proposed in Sec. \ref{DivBasedMethod} fail to significantly outperform baselines.

\begin{figure}[ht]
    \centering \includegraphics[width=0.5\textwidth]{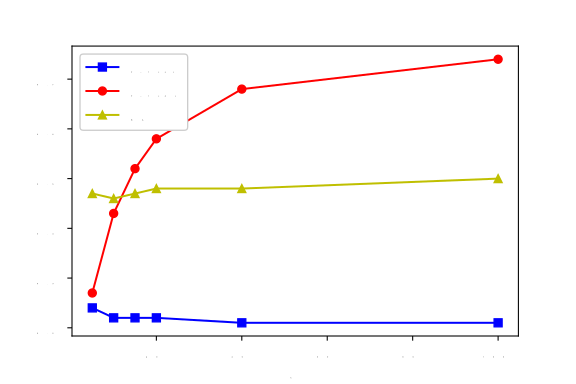}
    \caption{Unsupervised method, neighborhood-based SD, for ALL (mixed POS).}
    \label{fig:deltaSDneigh}
\end{figure}

In Figure \ref{fig:deltaSDneigh} we show the influence of $k$ in SD (neighbors) for the unsupervised $\Delta$ method. We see that while there is close to no change in balance accuracy, $F_1$ scores for both classes are more and more unbalanced as $k$ increases, indicating a more unfair model for high values of $k$. This is explained by the fact that the unsupervised model predicts more \textit{Diff} (dominant class) with higher $k$.

\subsection{Components of selected models}
\label{appendix:models}

Depending on the POS tag, the implementation strategies of \textit{SD}, \textit{DD} and frequency variables were different. Recall that these strategies were chosen given the average performances over 20 random train/test splits.

On adjectives, neighborhood based \textit{DD} as well as raw frequency counts was found to be better than alternatives. For \textit{SD}, cosine distance provides slightly higher performances than neighborhood-based measures, except when tuning the threshold for $\Delta$: in this case, SD(n$k$) with $k=15$ was best. Generally, for every method implying a neighbordhood based SD ($\Delta$(n$k$), LR multi., as well as the two non-linear models), a small/mid ranged $k$ was preferable (between 10 and 20).

For nouns, SD(cosine distance) was also the best choice except for $\Delta$ with tuned threshold: here, SD(n$k$) was preferred. Overall, the best range for the value of $k$ for neighbors-based SD was smaller (5 to 15). Frequency groups worked better than raw frequencies, while there was no difference in performance between the two definitions of DD.

Yet, for verbs, SD(n$k$) with $k=40$ actually outperforms cosine distance (except for unsupervised $\Delta$), and DD using Orthogonal Procrustes alignment and cosine distance \citep{hamilton-etal-2016-diachronic} was actually better than the definition relying on comparisons local neighborhoods. Both types of frequency variables (raw counts and groups) worked equally well. 

Finally, on the ALL dataset reuniting pairs accross POS tags, raw frequencies provide better results than groups. Cosine distance is better than neighborhoods for synchronic distances, and both techniques of diachronic distances performed similarly. For models forced to use SD(n$k$) in addition to SD(cd), the choice of $k$ did not really change the results.

\subsection{Predictive variables in our model}

In this supplementary section, we conduct a study about the role of some predictive variables in our best-performing Logistic Regression model, as potential sources of errors. The studied model uses SD with cosine-distance, both implementations of DD and raw frequency counts.

\label{analysisPred}

\begin{table}[ht]
    \centering
    \begin{tabular}{c||cc|cc|cc}\toprule
         & \multicolumn{2}{c|}{Pred.} & \multicolumn{2}{c|}{$y=$ Syn} & \multicolumn{2}{c}{$y=$ Diff} \\
         & Syn & Diff & TS & FD & TD & FS \\\cmidrule{1-7}
        $SD^{(T\firstTime)}$ & \textbf{.64} & \textbf{.83} & \textbf{.62} & \textbf{.84} & \textbf{.83} & \textbf{.64} \\
        $DD(\synOne)$ & .46 & .46 & .45 & .47 & .46 & .47 \\
        $DD(\synTwo)$ & \textbf{.50} & \textbf{.54} & \textbf{.48} & \textbf{.54} & \textbf{.54} & \textbf{.50} \\
        $FG^{(T2)}_\synOne$ & 2.3 & 2.2 & 2.4 & 2.1 & 2.2 & 2.2 \\
        $FG^{(T2)}_\synTwo$ & \textbf{1.9} & \textbf{1.4} & \textbf{2.1} & \textbf{1.5} & \textbf{1.4} & \textbf{1.8} \\
        
    \bottomrule
    \end{tabular}
    \caption{Average values of some variables for data subset based on the prediction of our best-performing LR model. TS,FS,TD,FD stand for True/False Syn/Diff. $FG_w^{(T)}$ stands for Frequency Groups of word $w$ at time $T$. Significant difference within a pair of columns are in bold.}
    \label{tab:comparisons}
\end{table}

For a selected number of variables, we look for significant differences between well-classified pairs and pairs with wrong prediction, in both classes separately. For a given variable, we estimate if a difference is significant between the well-classified and the misclassified samples of this class using a $t$-test for Gaussian distributed variables, or a Mann-Whitney $U$ test for other variables. A difference is significant if the $p$-value of the test is below 5\%. Results are reported in table \ref{tab:comparisons}. 

We observe significant differences of SD in pairs that are predicted as \textit{Syn} and those predicted as \textit{Diff} by our model, the first having a smaller SD at $T1$ than the latter. Because our model relies mostly on these SD to separate both classes, we wrongly classify \textit{Syn} pairs whose $SD^{(T\firstTime)}$ is close to that of \textit{Diff}, and conversely \textit{Diff} pairs whose $SD^{(T\firstTime)}$ is close to that of \textit{Syn} are misclassified. This indicates that our model still misses some subtleties that are now reflected by SD.

A similar non-separability of the distribution of "Syns" and "Diff" appears on DD and Frequency variable for the second word pair of the pair.
While it seems logical for our model to behave so regarding to the definition of LD, it is a clue that our input variables reflect noisy information that is confusing to the model. In the same idea, \citet{kutuzov-2022-contextualized} remarked that recent LSC detection models tend to raise False Positive, drawing attention to the limit of current models for LSC.

\end{document}